\documentclass[journal]{IEEEtran}
\usepackage{cite}
\usepackage{graphics}
\usepackage{graphicx}
\usepackage{caption}
\usepackage{amsfonts}
\usepackage{amsmath}
\usepackage{multirow}
\usepackage{gensymb}
\usepackage{subfigure}
\usepackage{palatino}
\usepackage{tikz}
\usepackage{latexsym}
\usetikzlibrary{shapes.geometric, arrows}

\begin{document}

\title{Unsupervised Lightweight Single Object Tracking with UHP-SOT++}

\author{Zhiruo~Zhou,~\IEEEmembership{Student~member,~IEEE,}
        Hongyu~Fu,~\IEEEmembership{Student~member,~IEEE,}
        Suya~You,
        and~C.-C.~Jay~Kuo,~\IEEEmembership{Fellow,~IEEE}
\thanks{Zhiruo Zhou, Hongyu Fu and C.-C. Jay Kuo are with the Ming-Hsieh
Department of Electrical and Computer Engineering, University of
Southern California, CA, 90089-2564, USA (e-mails: zhiruozh@usc.edu, 
hongyufu@usc.edu and cckuo@sipi.usc.edu).}%
\thanks{Suya You is with Army Research Laboratory, Adelphi,
Maryland, USA (e-mail: suya.you.civ@army.mil).}%
}

\maketitle

\begin{abstract}

An unsupervised, lightweight and high-performance single object tracker,
called UHP-SOT, was proposed by Zhou {\em et al.} recently. As an
extension, we present an enhanced version and name it UHP-SOT++ in this
work.  Built upon the foundation of the
discriminative-correlation-filters-based (DCF-based) tracker, two new
ingredients are introduced in UHP-SOT and UHP-SOT++: 1) background
motion modeling and 2) object box trajectory modeling.  The main
difference between UHP-SOT and UHP-SOT++ is the fusion strategy of
proposals from three models (i.e., DCF, background motion and object box
trajectory models). An improved fusion strategy is adopted by UHP-SOT++
for more robust tracking performance against large-scale tracking
datasets.  Our second contribution lies in an extensive evaluation of
the performance of state-of-the-art supervised and unsupervised methods
by testing them on four SOT benchmark datasets -- OTB2015, TC128, UAV123
and LaSOT. Experiments show that UHP-SOT++ outperforms all previous
unsupervised methods and several deep-learning (DL) methods in tracking
accuracy.  Since UHP-SOT++ has extremely small model size, high
tracking performance, and low computational complexity (operating at a
rate of 20 FPS on an i5 CPU even without code optimization), it is an
ideal solution in real-time object tracking on resource-limited
platforms. Based on the experimental results, we compare pros and cons
of supervised and unsupervised trackers and provide a new perspective to
understand the performance gap between supervised and unsupervised
methods, which is the third contribution of this work. 

\end{abstract}

\begin{IEEEkeywords}
Object tracking, online tracking, single object tracking, unsupervised tracking.
\end{IEEEkeywords}

\section{Introduction}\label{sec:introduction}

\IEEEPARstart{V}{ideo} object tracking is one of the fundamental
computer vision problems. It finds rich applications in video
surveillance \cite{xing2010multiple}, autonomous navigation
\cite{janai2020computer}, robotics vision \cite{zhang2015good}, etc.
Given a bounding box on the target object at the first frame, a tracker
has to predict object box locations and sizes for all remaining frames
in online single object tracking (SOT) \cite{yilmaz2006object}.  The
performance of a tracker is measured by accuracy (higher success rate),
robustness (automatic recovery from tracking loss), computational
complexity and speed (a higher number of frames per second of FPS).

Online trackers can be categorized into supervised and unsupervised ones
\cite{fiaz2019handcrafted}.  Supervised trackers based on deep learning
(DL) dominate the SOT field in recent years.  Some of them use a
pre-trained network such as AlexNet\cite{krizhevsky2012imagenet} or VGG
\cite{chatfield2014return} as the feature extractor and do online
tracking with extracted deep{} features \cite{danelljan2017eco,
danelljan2016beyond, ma2015hierarchical, qi2016hedged, wang2018multi}.
Others adopt an{} end-to-end optimized model which is trained by video
datasets in an offline manner \cite{li2018high, li2019siamrpn++} and
could be adapted to video frames in an online fashion \cite{lu2018deep,
nam2016learning, pu2018deep, song2017crest}.  The tracking problem is
formulated as a template matching problem in siamese trackers
\cite{bertinetto2016fully, li2018high, li2019siamrpn++, tao2016siamese,
zhu2018distractor, wang2018learning, he2018twofold}, which is popular
because of its simplicity and effectiveness. One recent trend is to
apply the Vision Transformer in visual tracking
\cite{wang2021transformer, chen2021transformer}. 

Although DL trackers offer state-of-the-art tracking accuracy,
they do have some limitations.  First, a large number of annotated
tracking video clips are needed in the training, which is a laborious
and costly task. Second, they demand large memory space to store the
parameters of deep networks due to large model sizes.  Third, the high
computational power requirement hinders their applications in
resource-limited devices such as drones or mobile phones. Fourth,
DL trackers need to be trained with video samples of diverse
content. Their capability in handling unseen objects appears to be
limited, which will be illustrated in the experimental section.  In
contrast with DL trackers, unsupervised trackers are attractive
since they do not need annotated boxes to train trackers. They are
favored in real-time tracking on resource-limited devices because of
lower power consumption. 

Advanced unsupervised SOT methods often use discriminative correlation
filters (DCFs). They were investigated between 2010 and 2018
\cite{bolme2010visual, henriques2014high, danelljan2015convolutional,
danelljan2016discriminative, danelljan2016beyond, bertinetto2016staple,
valmadre2017end, li2018learning}. DCF trackers conduct dense sampling
around the object box and solve a regression problem to learn a template
for similarity matching. Under the periodic sample assumption, matching
can be conducted very fast in the Fourier domain.  Spatial-temporal
regularized correlation filters (STRCF) \cite{li2018learning} adds
spatial-temporal regularization to template update and performs
favorably against other DCF trackers \cite{danelljan2017eco,
danelljan2015learning}. 

As deep neural networks (DNNs) get popular in recent
years, there is an increasing interest in learning DNN-based object
tracking models from offline videos without annotations. For example,
UDT+ \cite{wang2019unsupervised} and LUDT \cite{wang2021unsupervised}
investigated cycle learning in video, in which networks are trained to
track forward and backward with consistent object proposals. ResPUL
\cite{wu2021progressive} mined positive and negative samples from
unlabeled videos and leveraged them for supervised learning in building
spatial and temporal correspondence. These unsupervised deep trackers
reveal a promising direction in exploiting offline videos without
annotations. Yet, they are limited in performance. Furthermore, they
need the pre-training effort. In contrast, no pre-training on offline
datasets is needed in our unsupervised tracker.

Despite the above-mentioned developments in unsupervised trackers, there
is a significant performance gap between unsupervised DCF trackers and
supervised DL trackers. It is attributed to the limitations of DCF
trackers such as failure to recover from tracking loss and inflexibility
in object box adaptation. An unsupervised tracker, called UHP-SOT
(Unsupervised High-Performance Single Object Tracker), was recently
proposed in \cite{zhou2021uhp} to address the issues.  UHP-SOT used
STRCF as the baseline and incorporated two new modules -- background
motion modeling and trajectory-based object box prediction.  A simple
fusion rule was adopted by UHP-SOT to integrate proposals from three
modules into the final one.  UHP-SOT has the potential to recover from
tracking loss and offer flexibility in object box adaptation.  UHP-SOT
outperforms all previous unsupervised single object trackers and narrows
down the gap between unsupervised and supervised trackers.  It achieves
comparable performance against DL trackers on small-scale datasets such
as TB-50 and TB-100 (or OTB 2015) \cite{7001050}. 

This work is an extension of UHP-SOT with new contributions.  First, the
fusion strategies in UHP-SOT and UHP-SOT++ are different. The fusion
strategy in UHP-SOT was simple and ad hoc. UHP-SOT++ adopts a fusion
strategy that is more systematic and well justified. It is applicable to
both small- and large-scale datasets with more robust and accurate
performance.  Second, this work conducts more extensive experiments on
four object tracking benchmarks (i.e., OTB2015, TC128, UAV123 and LaSOT)
while only experimental results on OTB2015 were reported for UHP-SOT in
\cite{zhou2021uhp}. New experimental evaluations demonstrate that
UHP-SOT++ outperforms all previous unsupervised SOT methods (including
UHP-SOT) and achieves comparable results with DL methods on large-scale
datasets. Since UHP-SOT++ has an extremely small model size, high
tracking performance, and low computational complexity (operating at a
rate of 20 FPS on an i5 CPU even without code optimization), it is ideal
for real-time object tracking on resource-limited platforms. Third, we
make thorough discussion on pros and cons of supervised and unsupervised
trackers in this work. Besides quantitative evaluations, we provide a
few exemplary sequences with qualitative analysis on strengths and
weaknesses of UHP-SOT++ and its benchmarking methods. 

The rest of this paper is organized as follows. Related work is reviewed
in Sec. \ref{sec:review}. The UHP-SOT++ method is detailed in Sec.
\ref{sec:method}. Experimental results are shown in Sec.
\ref{sec:experiments}. Further discussion is provided in Sec.
\ref{sec:discussion}. Concluding remarks are given in Sec.
\ref{sec:conclusion}. 

\section{Related Work}\label{sec:review}

\begin{figure}[htbp]
\centerline{\includegraphics[width=\linewidth]{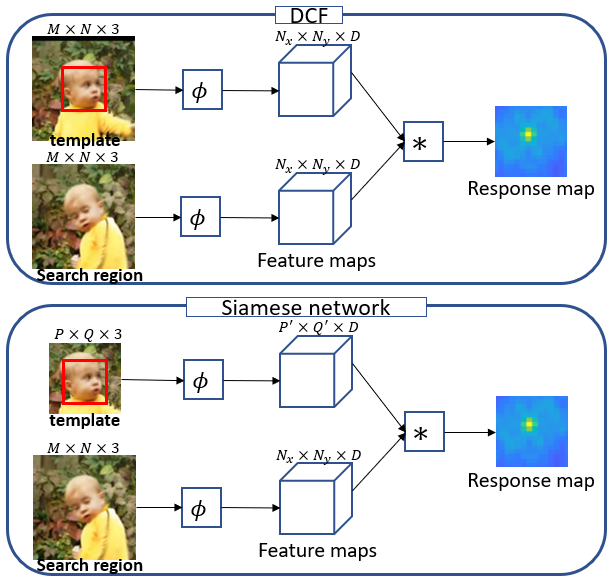}}
\caption{Comparison of the inference structures of a DCF tracker and a
siamese network tracker, where the red box denotes the object bounding
box, $\phi$ is the feature extraction module and $*$ is the correlation
operation.} \label{fig:dcfvssn}
\end{figure}

\subsection{DCF and Siamese Networks}\label{subsec:review}

Two representative single object trackers are reviewed and compared
here.  They are the DCF tracker and the siamese network tracker as shown
in Fig.~\ref{fig:dcfvssn}. The former is an unsupervised one while the
latter is a supervised one based on DL.  Both of them conduct template
matching within a search region to generate the response map for object
location.  The matched template in the next frame is centered at the
location that has the highest response. 

In the DCF, the template size is the same as that of the search region
so that the Fast Fourier Transform (FFT) could be used to speed up the
correlation process. 
To learn the template, a DCF uses the initial object patch to obtain a
linear template via regression in the Fourier domain:
\begin{equation}\label{eq:regression}
\mathrm{arg}\min_{\mathbf{f}} \frac{1}{2}\|\sum_{d=1}^D \mathbf{x}^{d}*
\mathbf{f}^{d}-\mathbf{y}\|^2,
\end{equation}
where $\mathbf{f}$ is the template to be determined, $\mathbf{x} \in
\mathbb{R}^{N_x \times N_y \times D}$ is the spatial map of $D$ features
extracted from the object patch, $*$ is the feature-wise spatial
convolution, and $\mathbf{y} \in \mathbb{R}^{N_x \times N_y}$ is a
centered Gaussian-shaped map that serves as the regression label. 
Templates in DCFs tend to contain some background information.
Furthermore, there exists boundary distortion caused by the 2D Fourier
transform. To alleviate these side effects, it is often to weigh the
template with a window function to suppress background and image
discontinuity. 

In contrast, the template size in a siamese networks is more flexible
and significantly smaller than the search region. Usually, the object
box of the first frame serves as the template for the search in all later frames. 
The correlation is typically implemented by the convolution which runs fast
on GPU. The shape and size of the predicted object bounding box are
determined by the regional proposal network inside the siamese network. 

\subsection{Spatial-Temporal Regularized Correlation Filters (STRCF)}
\label{subsec:STRCF}

STRCF is a DCF-based tracker. It has an improved regression objective
function using spatial-temporal regularization.  The template is
initialized at the first frame.  Suppose that the object appearance at
frame $t$ is modeled by a template, denoted by $\mathbf{f}_t$, which
will be used for similarity matching at frame $(t+1)$.  By modifying Eq.
(\ref{eq:regression}), STRCF updates its template at frame $t$ by
solving the following regression equation:
\begin{eqnarray}
\mathrm{arg}\min_{\mathbf{f}} & \Big\{ & \frac{1}{2}\|\sum_{d=1}^D \mathbf{x}_{t}^{d}*
\mathbf{f}^{d}-\mathbf{y}\|^2 +  \frac{1}{2}\sum_{d=1}^D \|\mathbf{w}\cdot 
\mathbf{f}^{d} \|^2 \nonumber \\
&& + \frac{\mu}{2}\|\mathbf{f}-\mathbf{f}_{t-1} \|^2, \Big\} \label{eq:strcf}
\end{eqnarray}
where $\mathbf{w}$ is the spatial weight on the template,
$\mathbf{f}_{t-1}$ is the template obtained from time $t-1$, and $\mu$
is a constant regularization coefficient.  We can interpret the three
terms in Eq. (\ref{eq:strcf}) as follows.  The first term is the
standard regression objective function of a DCF. The second term imposes
the spatial regularization. It gives more weights to features in the
center region of a template in the matching process.  The third term
imposes temporal regularization for smooth appearance change. 

To search for the box in frame $(t+1)$, STRCF correlates template
$\mathbf{f}_t$ with the search region and determines the new box
location by finding the location that gives the highest response.
Although STRCF can model the appearance change for general sequences, it
suffers from overfitting.  That is, it is not able to adapt to largely
deformed objects quickly.  Furthermore, it cannot recover from tracking
loss. The template model, $\mathbf{f}$, is updated at every frame with a
fixed regularization coefficient, $\mu$, in standard STRCF. 

Our UHP-SOT++ adopts STRCF as a building module. To address the
above-mentioned shortcomings, we have some modification in our
implementation. First, we skip updating $\mathbf{f}$ if no obvious
motion is observed. Second, a smaller $\mu$ is used when all modules
agree with each other in prediction so that $\mathbf{f}$ can adapt to
the new appearance of largely deformed objects faster. 

\section{Proposed UHP-SOT++ Method}\label{sec:method}

\begin{figure*}[htbp]
\centerline{\includegraphics[width=\textwidth]{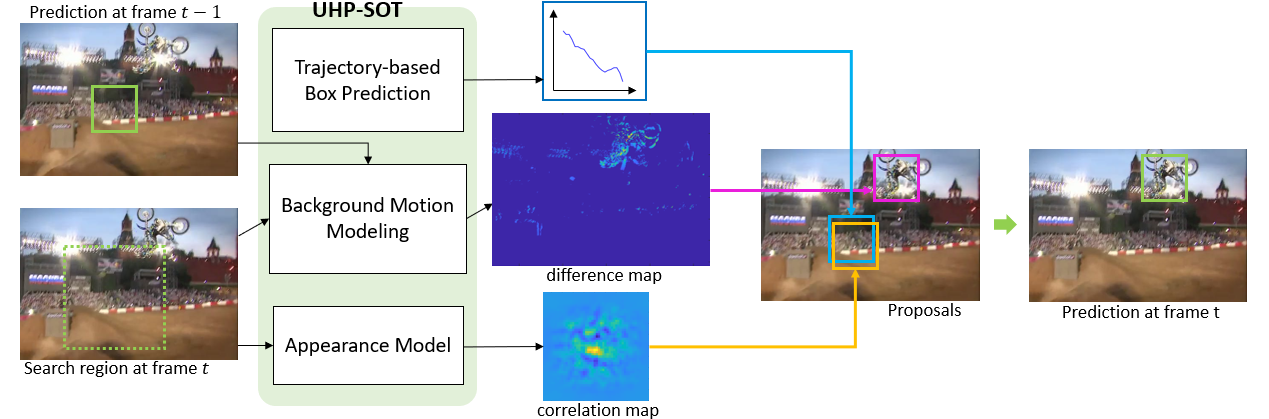}}
\caption{The system diagram of the proposed UHP-SOT++ method. It shows
one example where the object was lost at time $t-1$ but gets retrieved
at time $t$ because the proposal from background motion modeling is
accepted.} \label{fig:system}
\end{figure*}

\subsection{System Overview}\label{subsec:overview}

There are three main challenges in SOT: 
\begin{enumerate}
\item significant change of object appearance, 
\item loss of tracking,
\item rapid variation of object's location and/or shape.
\end{enumerate}
We propose a new tracker, UHP-SOT++, to address these challenges,
As shown in Fig. \ref{fig:system}, it consists of three modules:
\begin{enumerate}
\item appearance model update, 
\item background motion modeling,
\item trajectory-based box prediction.
\end{enumerate}

UHP-SOT++ follows the classic tracking-by-detection paradigm where the
object is detected within a region centered at its last predicted
location at each frame.  The histogram of gradients (HOG) features as
well as the color name (CN) \cite{danelljan2014adaptive} features are
extracted to yield the feature map. We choose the STRCF tracker
\cite{li2018learning} as the baseline because of its efficient and
effective appearance modeling and update. Yet, STRCF cannot handle the
second and the third challenges well because it only focuses on the
modeling of object appearance which could vary a lot across different
frames. Generally, the high variety of object appearance is difficult to
capture using a single model. Thus, we propose the second and the third
modules in UHP-SOT++ to enhance its tracking accuracy.  UHP-SOT++
operates in the following fashion.  The baseline tracker gets
initialized at the first frame. For the following frames, UHP-SOT++ gets
proposals from all three modules and merges them into the final
prediction based on a fusion strategy.  

The STRCF tracker was already discussed in Sec. \ref{subsec:STRCF}. For
the rest of this section, we will examine the background motion modeling
module and the trajectory-based box prediction module in Secs.
\ref{subsec:background} and \ref{subsec:trajectory}, respectively.
Finally, we will elaborate on the fusion strategy in Sec.
\ref{subsec:fusion}. Note that the fusion strategies of UHP-SOT and
UHP-SOT++ are completely different.

\subsection{Background Motion Modeling}\label{subsec:background}

We decompose the pixel displacement between adjacent frames (also called
optical flow) into two types: object motion and background motion.
Background motion is usually simpler, and it may be fit by a parametric
model.  Background motion estimation \cite{hariharakrishnan2005fast,
aggarwal2006object} finds applications in video stabilization, coding
and visual tracking. Here, we propose a 6-parameter model in form of
\begin{eqnarray}
x_{t+1}& = & \alpha_1 x_t + \alpha_2 y_t + \alpha_0, \\
y_{t+1}& = & \beta_1 x_t + \beta_2 y_t + \beta_0, 
\end{eqnarray}
where $(x_{t+1},y_{t+1})$ and $(x_t,y_t)$ are corresponding background
points in frames $(t+1)$ and $t$, respectively, and $\alpha_i$ and
$\beta_i$, $i=0,1,2$ are model parameters. With more than three pairs of
corresponding points, we can determine the model parameters using the
linear least-squares method. Usually, we choose a few salient points
(e.g., corners) to build the correspondence. We apply the background
model to the grayscale image $I_t(x,y)$ of frame $t$ to find the
estimated $\hat{I}_{t+1}(x,y)$ of frame $(t+1)$.  Then, we can compute
the difference map $\Delta I$:
\begin{equation}
\Delta I = \hat{I}_{t+1}(x,y) - I_{t+1}(x,y),
\end{equation}
which is expected to have small and large absolute values in the background and
foreground regions, respectively. Thus, we can determine potential
object locations.  While DCF trackers exploit foreground correlation to
locate the object, background modeling uses background correlation to
eliminate background influence in object tracking. They complement each
other. DCF trackers cannot recover from tracking loss easily since it
does not have a global view of the scene. In contrast, our background
modeling can find potential object locations by removing the background. 

\subsection{Trajectory-based Box Prediction}\label{subsec:trajectory}

\begin{figure}[!htbp]
\centerline{\includegraphics[width=\linewidth]{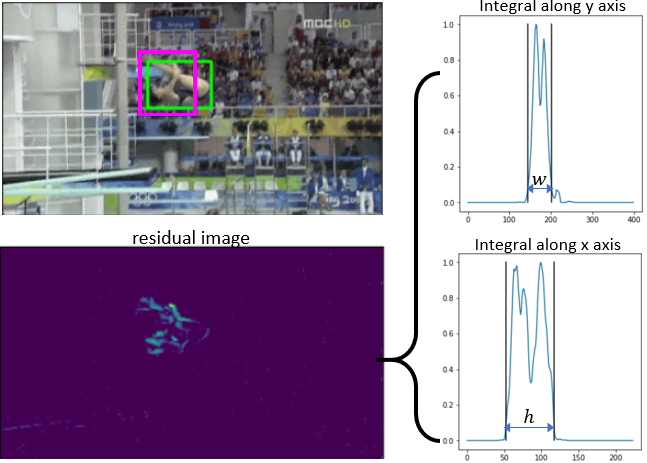}}
\caption{Illustration of shape change estimation based on background
motion model and trajectory-based box prediction, where the ground truth
and our proposal are annotated in green and magenta, respectively.}
\label{fig:shape}
\end{figure}

Given predicted box centers of the object of the last $N$ frames, $\{
(x_{t-N},y_{t-N}),\cdots,(x_{t-1},y_{t-1}) \}$, we calculate $N-1$
displacement vectors $\{ (\Delta x_{t-N+1},\Delta y_{t-N+1}),\cdots,(\Delta
x_{t-1},\Delta y_{t-1}) \}$ and apply the principal component analysis
(PCA) to them.  To predict the displacement at frame $t$, we fit the
first principal component using a line and set the second principal
component to zero to remove noise.  Then, the center location of the
box at frame $t$ can be written as
\begin{equation}
(\hat{x}_{t}, \hat{y}_{t}) = (x_{t-1}, y_{t-1}) + (\hat{\Delta x}_{t}, 
\hat{\Delta y}_{t}).
\end{equation}
Similarly, we can estimate the width and the height of the box at frame
$t$, denoted by $(\hat{w}_t, \hat{h}_t)$.  Typically, the physical
motion of an object has an inertia in motion trajectory and its size,
and the box prediction process attempts to maintain the inertia. It
contributes to better tracking performance in two ways.  First, it
removes small fluctuation of the box in its location and size. Second,
when there is a rapid deformation of the target object, the appearance
model alone cannot capture the shape change effectively. In contrast,
the combination of background motion modeling and the trajectory-based
box prediction can offer a more satisfactory solution. For example, Fig.
\ref{fig:shape}, shows a frame of the \textit{diving} sequence in the
upper-left subfigure, where the green and the magenta boxes are the ground
truth and the result of UHP-SOT++, respectively.  Although a DCF tracker
can detect the size change by comparing correlation scores at five image
resolutions, it cannot estimate the aspect ratio change properly.  In
contrast, as shown in the lower-left subfigure, the residual image after
background removal in UHP-SOT++ reveals the object shape. By summing up
absolute pixel values of the residual image horizontally and vertically
and using a threshold to determine two ends of the box, we have
\begin{equation}
\hat{w}=x_{\max}-x_{\min}, \mbox{  and  } \hat{h}=y_{\max}-y_{\min}. 
\end{equation}
Note that raw estimates may not be stable across different frames.
Estimates that deviate much from the trajectory of $(\Delta
w_{t},\Delta h_{t})$ are rejected to yield a robust and
deformable box proposal. 

\begin{figure*}[htbp]
\centerline{\includegraphics[width=\textwidth]{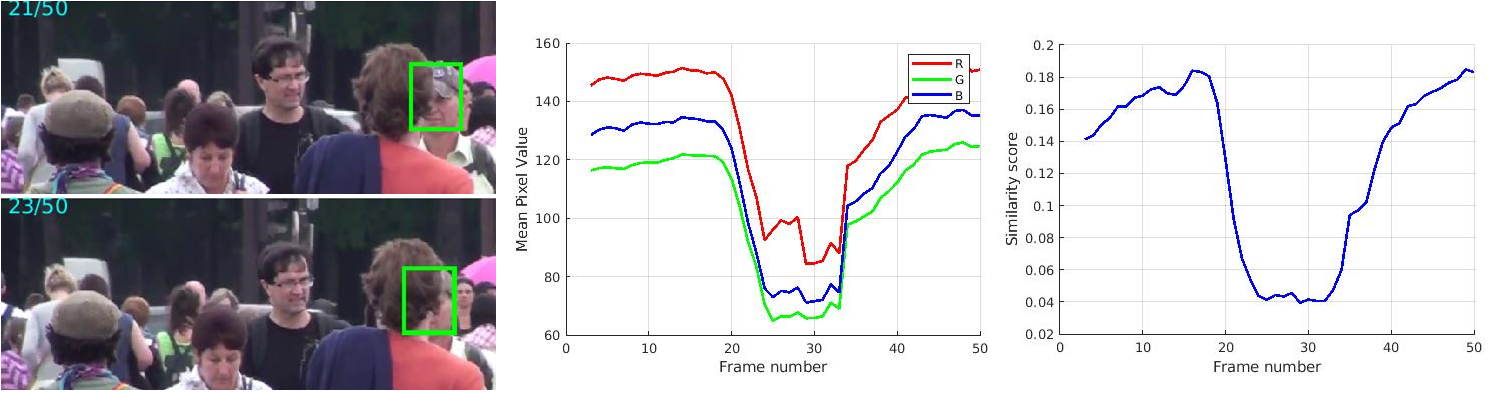}}
\caption{Illustration of occlusion detection, where the green box shows
the object location. The color information and similarity score could
change rapidly if occlusion occurs.} \label{fig:fusion_occ}
\end{figure*}

\begin{table*}[htbp]
\caption{All tracking scenarios are classified into 8 cases in terms of
the overall quality of proposals from three modules. The fusion strategy
is set up for each scenario. The update rate is related the
regularization coefficient, $\mu$, that controls to which extent the
appearance model should be updated.}\label{tab3:fusion}
\begin{center}
\begin{tabular}{ccccc} \hline
 $isGood_{app}$ & $isGood_{trj}$ & $isGood_{bgd}$ & Proposal to take & Update rate \\ \hline \hline
1 & 1 & 1 & $B_\mathrm{app}$ or union of three & normal \\
1 & 1 & 0 & $B_\mathrm{app}$ or $B_\mathrm{trj}$ or union of two & normal \\
1 & 0 & 1 & $B_\mathrm{app}$ or $B_\mathrm{bgd}$ or union of two & normal \\
0 & 1 & 1 & $B_\mathrm{trj}$ or $B_\mathrm{bgd}$ or union of two & normal or stronger \\
1 & 0 & 0 & $B_\mathrm{app}$ & normal \\
0 & 1 & 0 & $B_\mathrm{app}$ or $B_\mathrm{trj}$ & normal or stronger \\
0 & 0 & 1 & $B_\mathrm{app}$ or $B_\mathrm{bgd}$ & normal or stronger \\
0 & 0 & 0 & $B_\mathrm{app}$ or last prediction in case of occlusion & normal or weaker \\
\hline
\end{tabular}
\end{center}
\end{table*}

\begin{figure*}[htbp]
\centerline{\includegraphics[width=\textwidth]{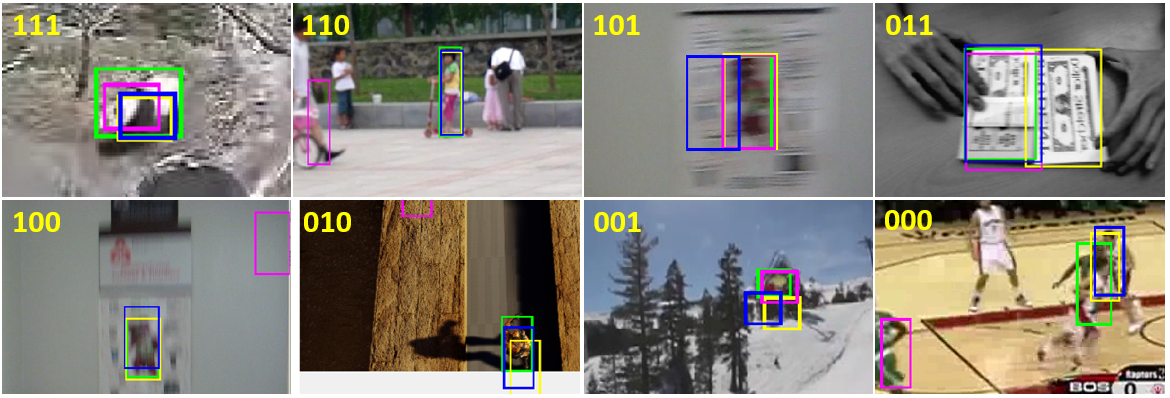}}
\caption{An example of quality assessment of proposals, where the green box is the
ground truth, and yellow, blue and magenta boxes are proposals from
$B_\mathrm{app}$, $B_\mathrm{trj}$ and $B_\mathrm{bgd}$, respectively,
and the bright yellow text on the top-left corner denotes the quality of
three proposals $(isGood_{app},isGood_{trj}, isGood_{bgd})$.}
\label{fig:fusion_showcase}
\end{figure*}

\subsection{Fusion Strategy}\label{subsec:fusion}

We have three box proposals for the target object at frame $t$: 1)
$B_\mathrm{app}$ from the baseline STRCF tracker to capture appearance
change, 2) $B_\mathrm{bgd}$ from the background motion predictor to
eliminate unlikely object regions, and 3) $B_\mathrm{trj}$ from the
trajectory predictor to maintain the inertia of the box position and
size. A fusion strategy is needed to yield the final box location and
size. We consider a couple of factors for its design.

\subsubsection{Proposal Quality} There are three box proposals. The
quality of each box proposal can be measured by: 1) object appearance
similarity, and 2) robustness against the trajectory.  We use a binary
flag to indicate whether the quality of a proposal is good or not.  As
shown in Table \ref{tab3:fusion}, the flag is set to one if a proposal
keeps proper appearance similarity and is robust against trajectory.
Otherwise, it is set to zero. 

For the first measure, we store two appearance models: the latest model,
$\mathbf{f}_{t-1}$, and an older model, $\mathbf{f}_{i}$, $i\leq t-1$,
where $i$ is the last time instance where all three boxes have the same
location.  Model $\mathbf{f}_{i}$ is less likely to be contaminated
since it needs agreement from all modules. To check the reliability of
the three proposals, we compute correlation scores for the following six
pairs: ($\mathbf{f}_{t-1}$, $B_\mathrm{app}$), ($\mathbf{f}_{t-1}$,
$B_\mathrm{trj}$), ($\mathbf{f}_{t-1}$, $B_\mathrm{bgd}$),
($\mathbf{f}_{i}$, $B_\mathrm{app}$), ($\mathbf{f}_{i}$,
$B_\mathrm{trj}$), and ($\mathbf{f}_{i}$, $B_\mathrm{bgd}$). They
provide appearance similarity measures of the two previous models
against the current three proposals. A proposal has good similarity if
one of its correlation scores is higher than a threshold. 

For the second measure, if $B_\mathrm{app}$ and $B_\mathrm{trj}$ have a
small displacement (say, 30 pixels) from the last prediction, the move
is robust. As to $B_\mathrm{bgd}$, it often jumps around and, thus, is
less reliable. However, if the standard deviations of its historical
locations along the $x$-axis and $y$-axis are small enough (e.g., 30
pixels over the past 10 frames), then they are reliable. 

\subsubsection{Occlusion Detection} We propose an occlusion detection
strategy for color images, which is illustrated in
Fig.~\ref{fig:fusion_occ}. As occlusion occurs, we often observe a
sudden drop in the similarity score and a rapid change on the averaged
RGB color values inside the box. A drop is sudden if the mean over the
past several frames is high while the current value is significantly
lower. If this is detected, we keep the new prediction the same as the
last predicted position since the new prediction is unreliable. We do
not update the model for this frame either to avoid drifting and/or
contamination of the appearance model. 

\subsubsection{Rule-based Fusion} Since each of the three proposals has
a binary flag, all tracking scenarios can be categorized into 8 cases as
shown in Fig.~\ref{fig:fusion_showcase}. We propose a fusion scheme for
each case below. 
\begin{itemize}
\item When all three proposals are good, their boxes are merged together as a
minimum covering rectangle if they overlap with each other with IoU above a threshold.
Otherwise, $B_\mathrm{app}$ is adopted. 
\item When two proposals are good, merge them if they overlap with each other 
with IoU above a threshold. Otherwise, the one with better robustness is adopted.
\item When one proposal is good, adopt that one if it is
$B_\mathrm{app}$.  Otherwise, that proposal is compared with $B_\mathrm{app}$ to 
verify its superiority by observing a higher similarity score or better
robustness.
\item When all proposals have poor quality, the occlusion detection process is conducted. 
The last prediction is adopted in case of occlusion. Otherwise, $B_\mathrm{app}$ is adopted.
\item When other proposals outperform $B_\mathrm{app}$, the
regularization coefficient, $\mu$, is adjusted accordingly for stronger
update. Because this might reveal that the appearance model needs to be
updated more to capture the new appearance. 
\end{itemize}
The fusion rule is summarized in Table \ref{tab3:fusion}. In most cases,
$B_\mathrm{app}$ is reliable and it will be chosen or merged with other
proposals because the change is smooth between adjacent frames in the
great majority of frames in a video clip. 

\section{Experiments}\label{sec:experiments}

\subsection{Experimental Set-up}\label{subsec:setup}

To show the performance of UHP-SOT++, we compare it with several
state-of-the-art unsupervised and supervised trackers on four single
object tracking datasets. They are OTB2015 \cite{7001050}, TC128
\cite{liang2015encoding}, UAV123 \cite{mueller2016benchmark} and LaSOT
\cite{fan2019lasot}. OTB2015 (also named OTB in short) and TC128, which
contain 100 and 128 color or grayscale video sequences, respectively,
are two widely used small-scale datasets.  UAV123 is a larger one, which
has 123 video sequences with more than 110K frames in total. Videos in
UAV123 are captured by low-altitude drones. They are useful in the
tracking test of small objects with a rapid change of viewpoints. LaSOT
is the largest single object tracking dataset that targets at
diversified object classes and flexible motion trajectories in longer
sequences. It has one training set with dense annotation for supervised
trackers to learn and another test set for performance evaluation. The
test set contains 280 videos of around 685K frames. 

Performance evaluation is conducted using the ``One Pass Evaluation
(OPE)" protocol.  The metrics include the precision plot (i.e., the
distance of the predicted and actual box centers) and the success plot
(i.e., overlapping ratios at various thresholds). The distance precision (DP)
is measured at the 20-pixel threshold to rank different methods. The
overlap precision is measured by the area-under-curve (AUC) score.  We
use the same hyperparameters as those in STRCF except for regularization
coefficient, $\mu$.  If the appearance box is not chosen, STRCF sets
$\mu=15$ while UHP-SOT++ selects $\mu \in\{15,10,5,0\}$. The smaller $\mu$ is,
the stronger the update is. The number of previous frames for
trajectory prediction is $N=20$.  The cutting threshold along the
horizontal or vertical direction is set 0.1. The threshold for 
good similarity score is 0.08, and a threshold of 0.5 for IoU is adopted. 
UHP-SOT++ runs at 20 frames
per second (FPS) on a PC equipped with an Intel(R) Core(TM) i5-9400F
CPU.  The speed data of other trackers are either from their original
papers or benchmarks. Since no code optimization is
conducted, all reported speed data should be viewed as lower bounds for
the corresponding trackers.

\subsection{Ablation study}

\begin{figure}[htbp]
\centerline{\includegraphics[width=\linewidth]{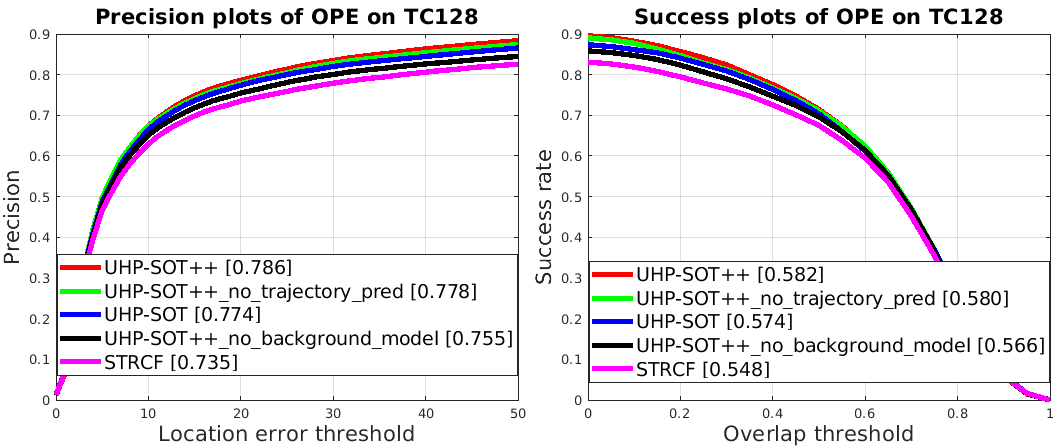}}
\caption{The precision plot and the success plot 
of our UHP-SOT++ tracker with different configurations on the TC128 
dataset, where the numbers inside the parentheses are the DP values 
and AUC scores, respectively.} \label{fig:ablation_study}
\end{figure}

We compare different configurations of UHP-SOT++ on
the TC128 dataset to investigate contributions from each module in
Fig.~\ref{fig:ablation_study}. As compared with UHP-SOT, improvements on
both DP and AUC in UHP-SOT++ come from the new fusion strategy. Under
this strategy, the background motion modeling plays an more important
role and it has comparable performance even without the trajectory
prediction.  Although the trajectory prediction module is simple, it
contributes a lot to higher tracking accuracy and robustness as revealed
by the performance improvement over the baseline STRCF.

\begin{figure}[htbp]
\centerline{\includegraphics[width=\linewidth]{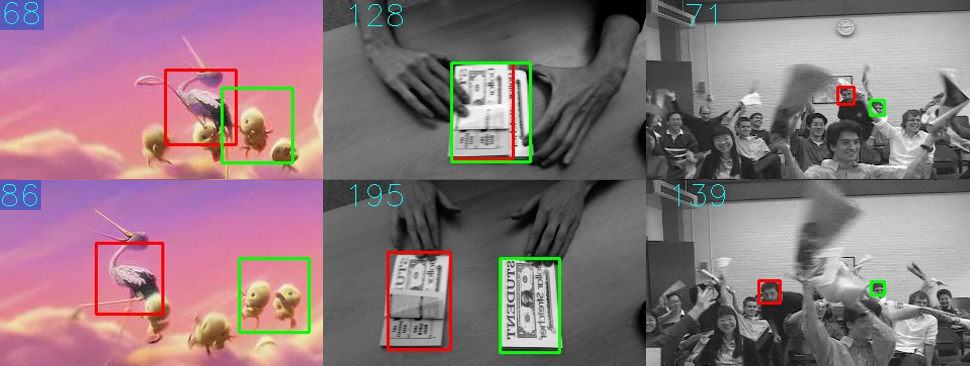}}
\caption{Failure cases of UHP-SOT++ (in green)
as compared to UHP-SOT (in red) on OTB2015.} \label{fig:analysis_error}
\end{figure}

More performance comparison between UHP-SOT++,
UHP-SOT and STRCF is presented in Table \ref{tab:sota}. As compared with
STRCF, UHP-SOT++ achieves 1.8\%, 6.2\%, 6.7\% and 6.8\% gains in the
success rate on OTB, TC128, UAV123 and LaSOT, respectively. As to the
mean precision, it has an improvement of 1.2\%, 6.9\%, 7.2\% and 10.4\%,
respectively.  Except for OTB, UHP-SOT++ outperforms UHP-SOT in both the
success rate and the precision. This is especially obvious for
large-scale datasets.  Generally, UHP-SOT++ has better tracking
capability than UHP-SOT. Its performance drop in OTB is due to the
tracking loss in three sequences; namely, \textit{Bird2},
\textit{Coupon} and \textit{Freeman4}. They have complicated appearance
changes such as severe rotation, background clutter and heavy occlusion.
As shown in Fig.~\ref{fig:analysis_error}, errors at some key frames
lead to total loss of the object, and the lost object cannot be easily
recovered from motion. The trivial fusion strategy based on appearance
similarity in UHP-SOT seems to work well on their key frames while the
fusion strategy of UHP-SOT++ does not suppress wrong proposals properly
since background clutters have stable motion and trajectories as well.

\begin{table*}[htbp]
\caption{Comparison of state-of-the-art supervised
and unsupervised trackers on four datasets, where the performance is
measure by the distance precision (DP) and the area-under-curve (AUC) score in
percentage. The model size is measured by the memory required to store
needed data such as the model parameters of pre-trained networks. The
best unsupervised performance is highlighted. Also, S and P indicate
\textbf{S}upervised and \textbf{P}re-trained, respectively.}\label{tab:sota}
\begin{center}
\begin{tabular}{cccc||cc|cc|cc|cc||ccc} \hline
Trackers & Year & \textbf{S} & \textbf{P} & \multicolumn{2}{c}{OTB2015} & \multicolumn{2}{c}{TC128} & \multicolumn{2}{c}{UAV123} & \multicolumn{2}{c||}{LaSOT} & FPS & Device & Model size (MB)\\
 & & & & DP & AUC & DP & AUC & DP & AUC & DP & AUC & & & \\
\hline
SiamRPN++\cite{li2019siamrpn++} & 2019 & \checkmark & \checkmark & 91.0&69.2 & -&- & 84.0&64.2 & 49.3&49.5 & 35 & GPU & 206 \\
ECO\cite{danelljan2017eco} & 2017 & \checkmark & \checkmark & 90.0&68.6 & 80.0&59.7 & 74.1&52.5 & 30.1&32.4 & 10 & GPU & 329 \\
\hline
UDT+\cite{wang2019unsupervised} & 2019 & $\times$ & \checkmark & 83.1&63.2 & 71.7&54.1 & -&- & -&- & 55 & GPU & $<1$\\
LUDT\cite{wang2021unsupervised} & 2020 & $\times$ & \checkmark & 76.9&60.2 & 67.1&51.5 & -&- & -&26.2 & 70 & GPU & $<1$\\
ResPUL\cite{wu2021progressive} & 2021 & $\times$ & \checkmark & -&58.4 & -&- & -&- & -&- & - & GPU & $>6$\\
\hline
ECO-HC\cite{danelljan2017eco} & 2017 & $\times$ & $\times$ & 85.0&63.8 & 75.3&55.1 & 72.5&50.6 & 27.9&30.4 & 42 & CPU & $<1$\\
STRCF\cite{li2018learning} & 2018 & $\times$ & $\times$ & 86.6&65.8 & 73.5&54.8 & 67.8&47.8 & 29.8&30.8 & 24 & CPU & $<1$\\
UHP-SOT\cite{zhou2021uhp} & 2021 & $\times$ & $\times$ & \textbf{90.9}&\textbf{68.9} & 77.4&57.4 & 71.0&50.1 & 31.1&32.0 & 23 & CPU & $<1$\\
UHP-SOT++ & Ours & $\times$ & $\times$ & 87.6&66.9 & \textbf{78.6}&\textbf{58.2} & \textbf{72.7}&\textbf{51.0} & \textbf{32.9}&\textbf{32.9} & 20 & CPU & $<1$\\
\hline
\end{tabular}
\end{center}
\end{table*}

\begin{figure}[htbp]
\centerline{\includegraphics[width=\linewidth]{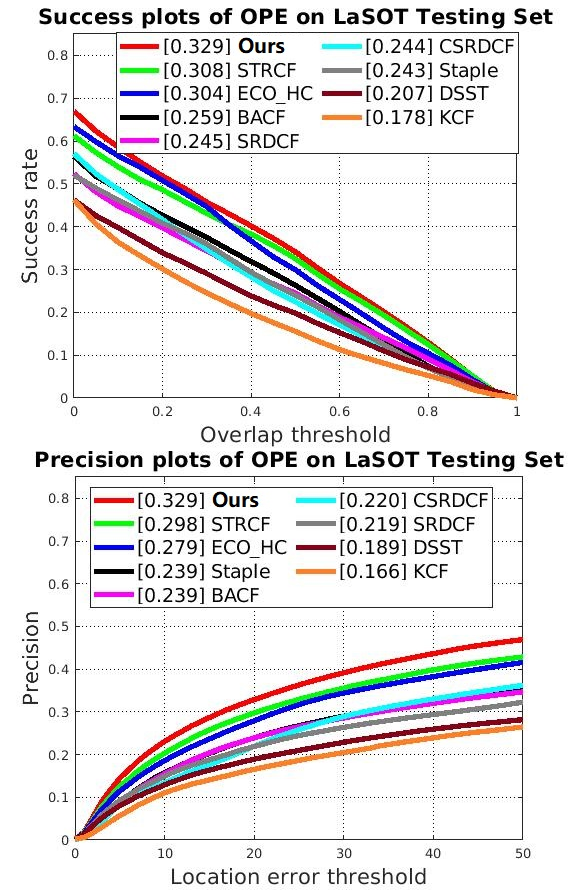}}
\caption{The success plot and the precision plot of nine unsupervised
tracking methods for the LaSOT dataset, where the numbers inside the
parentheses are the overlap precision and the distance precision values,
respectively.} \label{fig:comp_benchmark_un}
\end{figure}

\begin{figure}[htbp]
\centerline{\includegraphics[width=\linewidth]{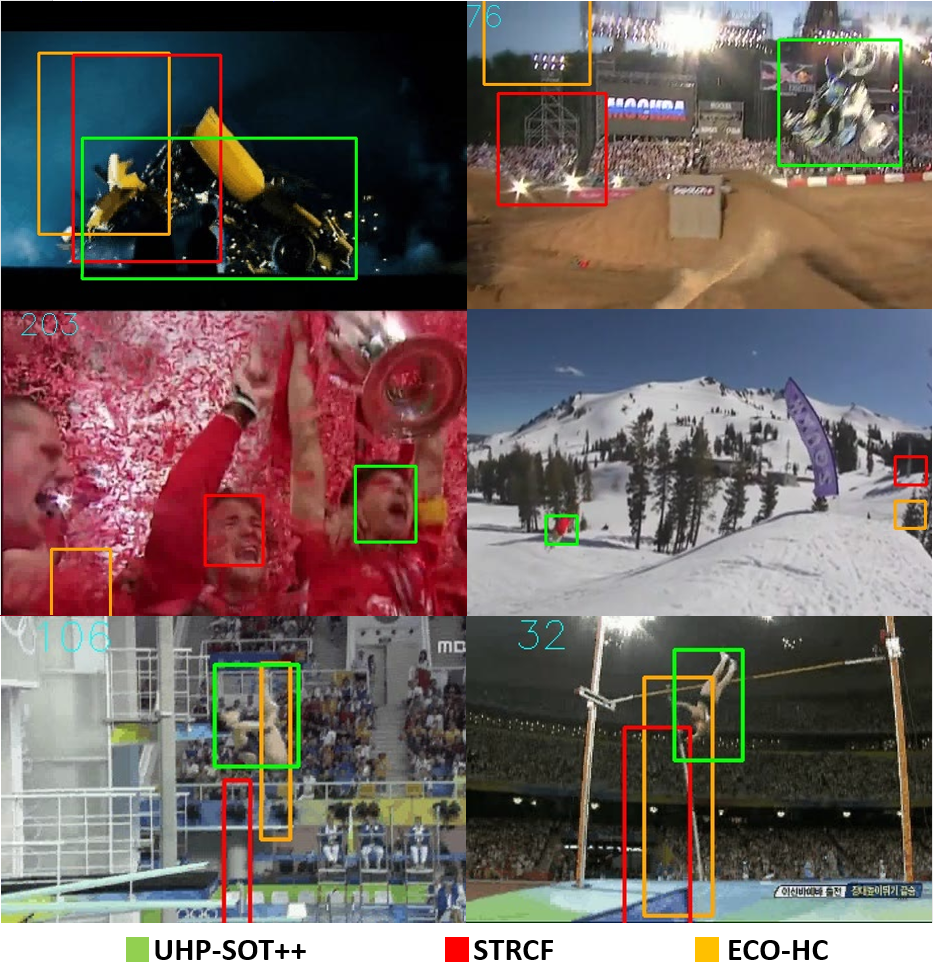}}
\caption{Qualitative evaluation of three leading unsupervised trackers,
where UHP-SOT++ offers a robust and flexible box prediction.}
\label{fig:showcase_unsupervised}
\end{figure}

\begin{figure*}[htbp]
\centerline{\includegraphics[width=\textwidth]{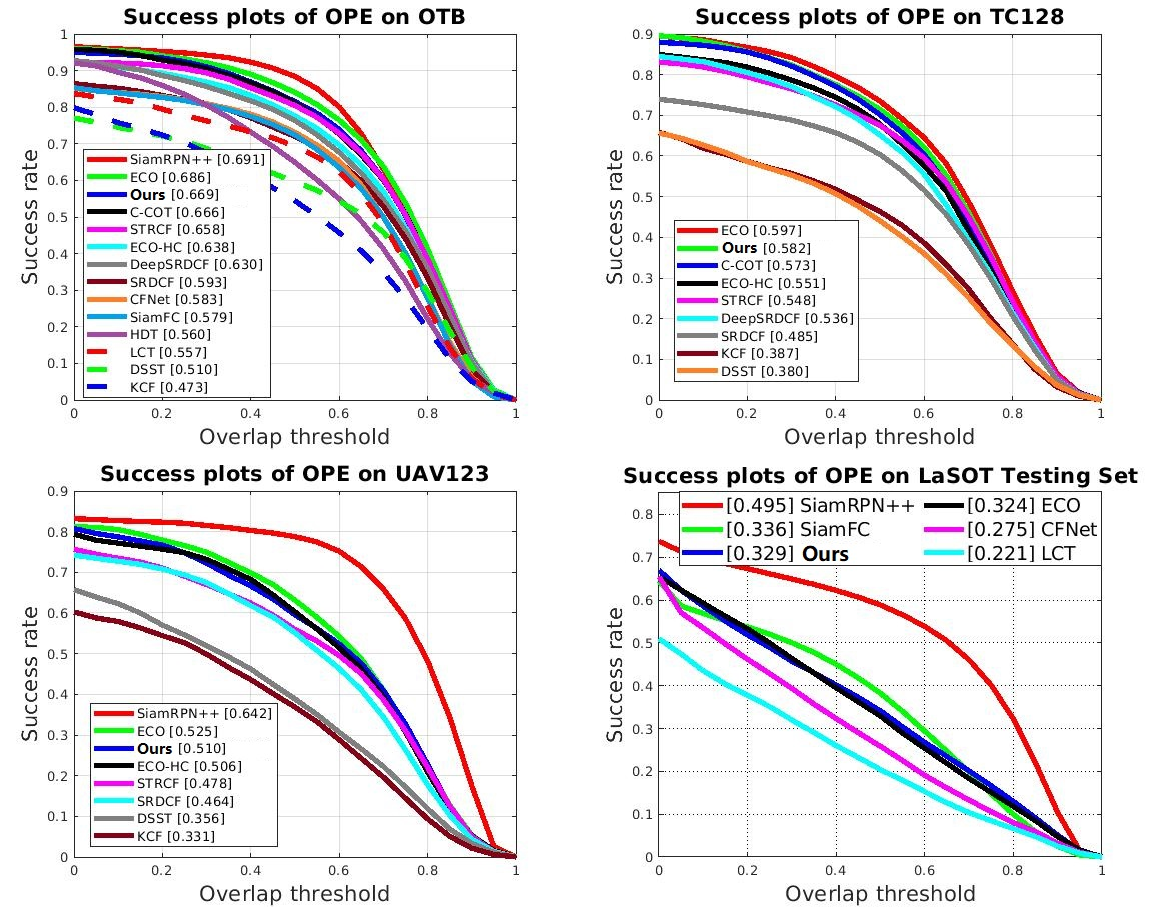}}
\caption{The success plot comparison of UHP-SOT++ with several
supervised and unsupervised tracking methods on four datasets, where
only trackers with raw results published by authors are listed.  For the
LaSOT dataset, only supervised trackers are included for performance
benchmarking in the plot since the success plot of unsupervised methods
is already given in Fig.~\ref{fig:comp_benchmark_un}.}
\label{fig:comp_benchmark}
\end{figure*}

\subsection{Comparison with State-of-the-art Trackers}

We compare the performance of UHP-SOT++ and several unsupervised
trackers for the LaSOT dataset in Fig. \ref{fig:comp_benchmark_un}. The
list of benchmarking methods includes ECO-HC \cite{danelljan2017eco},
STRCF \cite{li2018learning}, CSR-DCF \cite{alan2018discriminative},
SRDCF \cite{danelljan2015learning}, Staple \cite{bertinetto2016staple},
KCF \cite{henriques2014high}, DSST \cite{danelljan2016discriminative}.
UHP-SOT++ outperforms other unsupervised methods by a large margin,
which is larger than 0.02 in the mean scores of the success rate and the
precision.  Besides, its running speed is 20 FPS, which is comparable
with that of the second runner STRCF (24 FPS) and the third runner (42
FPS) based on experiments in OTB. With a small increase in computational
and memory resources, UHP-SOT++ gains in tracking performance by adding
object box trajectory and background motion modeling modules.  Object
boxes of three leading unsupervised trackers are visualized in
Fig.~\ref{fig:showcase_unsupervised} for qualitative performance
comparison. As compared with other methods, the proposals of UHP-SOT++
offer a robust and flexible box prediction.  They follow tightly with
the object in both location and shape even under challenging scenarios
such as motion blur and rapid shape change. 

We compare the success rates of UHP-SOT++ and several supervised and
unsupervised trackers against all four datasets in Fig.
\ref{fig:comp_benchmark}. Note that there are more benchmarking methods
for OTB but fewer for TC128, UA123 and LaSOT since OTB is an earlier
dataset. The supervised deep trackers under consideration include SiamRPN++
\cite{li2019siamrpn++}, ECO \cite{danelljan2017eco}, C-COT
\cite{danelljan2016beyond}, DeepSRDCF \cite{danelljan2015learning}, HDT
\cite{fiaz2019handcrafted}, SiamFC\_3s \cite{bertinetto2016fully}, CFNet
\cite{valmadre2017end} and LCT \cite{ma2015long}. Other deep
trackers that have leading performance but are not likely to be used on
resource-limited devices due to their extremely high complexity, such as
transformer-based trackers \cite{wang2021transformer,
chen2021transformer}, are not included here. Although the
performance of a tracker may vary from one dataset to the other due to
different video sequences collected by each dataset, UHP-SOT++ is among
the top three in all four datasets.  This demonstrates the
generalization capability of UHP-SOT++. Its better performance than ECO
on LaSOT indicates a robust and effective update of the object model.
Otherwise, it would degrade quickly with worse performance because of
longer LaSOT sequences. Besides, its tracking speed of 20 FPS on
 CPU is faster than many deep trackers such as ECO (10 FPS),
DeepSRDCF (0.2 FPS), C-COT (0.8 FPS) and HDT (2.7 FPS). 

In Table \ref{tab:sota}, we further compare
UHP-SOT++ with state-of-the-art unsupervised deep trackers UDT+
\cite{wang2019unsupervised}, LUDT \cite{wang2021unsupervised} and ResPUL
\cite{wu2021progressive} in their AUC and DP values, running speeds and
model sizes. Two leading supervised trackers SiamRPN++ and ECO in Fig.
\ref{fig:comp_benchmark} are also included. It shows that UHP-SOT++
outperforms recent unsupervised deep trackers by a large margin. We
should emphasize that deep trackers demand pre-training on offline
datasets while UHP-SOT++ does not. In addition, UHP-SOT++ is attractive
because of its lower memory requirement and near real-time running speed
on CPUs.  Although ECO-HC also provides a light-weight solution, there
is a performance gap between UHP-SOT++ and ECO-HC.  SiamRPN++ has the
best tracking performance among all trackers, due to the merit of
end-to-end optimized network with auxiliary modules such as
classification head and the region proposal network. Yet, its large
model size and GPU hardware requirement limit its applicability in
resource-limited devices such as mobile phones or drones. In addition,
as an end-to-end optimized deep tracker, SiamRPN++ has the
interpretability issue to be discussed later. 

\subsection{Attribute-based Study}

\begin{figure*}[htbp]
\centerline{\includegraphics[width=\linewidth]{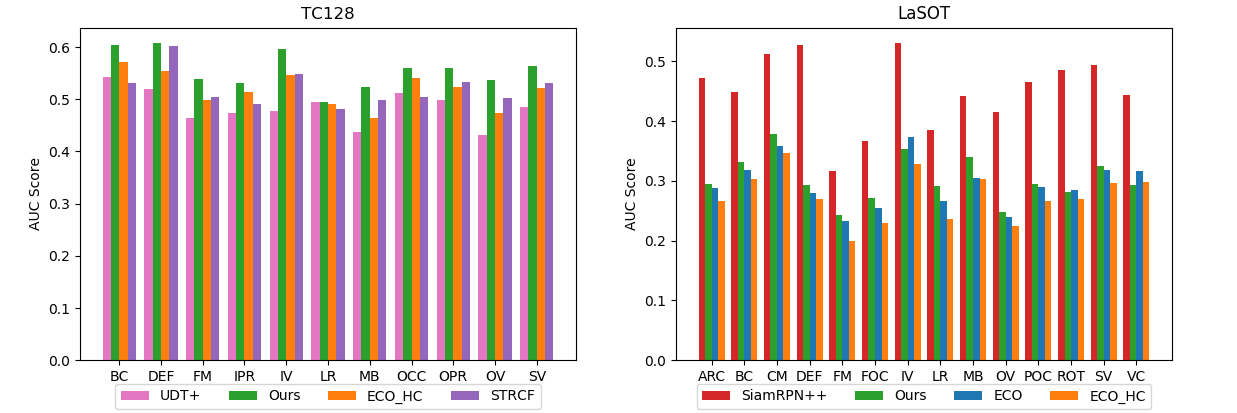}}
\caption{The area-under-curve (AUC) scores for two
datasets, TC128 and LaSOT, under the attribute-based evaluation, where
attributes of concern include the aspect ratio change (ARC), background
clutter (BC), camera motion (CM), deformation (DEF), fast motion (FM),
full occlusion (FOC), in-plane rotation (IPR), illumination variation
(IV), low resolution (LR), motion blur (MB), occlusion (OCC),
out-of-plane rotation (OPR), out-of-view (OV), partial occlusion (POC),
scale variation (SV) and viewpoint change (VC),
respectively.}\label{fig:attrs}
\end{figure*}

To better understand the capability of different trackers, we analyze
the performance variation under various challenging tracking conditions.
These conditions can be classified into the following attributes: aspect
ratio change (ARC), background clutter (BC), camera motion (CM),
deformation (DEF), fast motion (FM), full occlusion (FOC), in-plane
rotation (IPR), illumination variation (IV), low resolution (LR), motion
blur (MB), occlusion (OCC), out-of-plane rotation (OPR), out-of-view
(OV), partial occlusion (POC), scale variation (SV) and viewpoint change
(VC). We compare the AUC scores of supervised
trackers (e.g., SiamRPN++ and ECO) and unsupervised trackers (e.g.,
UHP-SOT++, ECO-HC, UDT+ and STRCF) under these attributes in Fig.
\ref{fig:attrs}.

We have the following observations. First, among
unsupervised trackers, UHP-SOT++ has leading performance in all
attributes, which reveals improved robustness from its basic modules and
fusion strategy. Second, although ECO utilizes deep features, it is
weak in flexible box regression and, as a result, it is outperformed by
UHP-SOT++ in handling such deformation and shape changes against LaSOT.
In contrast, SiamRPN++ is better than other trackers especially in DEF
(deformation), ROT (rotation)and VC (viewpoint change).  The superior
performance of SiamRPN++ demonstrates the power of its region proposal
network (RPN) in generating tight boxes. The RPN inside SiamRPN++ not
only improves IoU score but also has the long-term benefit by excluding
noisy information.  Fourth, supervised trackers perform better in IV
(illumination variation) and LR (low resolution) than unsupervised
trackers in general. This can be explained by the fact that unsupervised
trackers adopt HOG, CN features or other shallow features which do not
work well under these attributes.  They focus on local structures of the
appearance and tend to fail to capture the object when the local
gradient or color information is not stable.  Finally, even with the
feature limitations, UHP-SOT++ still runs second in many attributes
against LaSOT because of the stability offered by trajectory
prediction and its capability to recover from tracking loss via
background motion modeling. 

\begin{figure*}[htbp]
\centerline{\includegraphics[width=\linewidth]{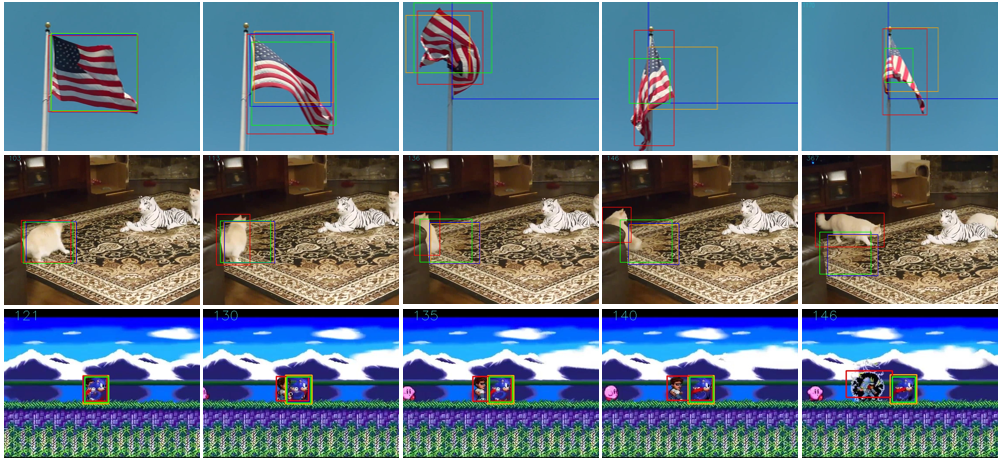}}
\caption{Qualitative comparison of top runners
against the LaSOT dataset, where tracking boxes of SiamRPN++, UHP-SOT++,
ECO and ECO-HC are shown in red, green, blue and yellow, respectively.
The first two rows show sequences in which SiamRPN++ outperforms others
significantly while the last row offers the sequence in which SiamRPN++
performs poorly.}\label{fig:error_showcase}
\end{figure*}

\begin{figure*}[htbp]
\centerline{\includegraphics[width=\linewidth]{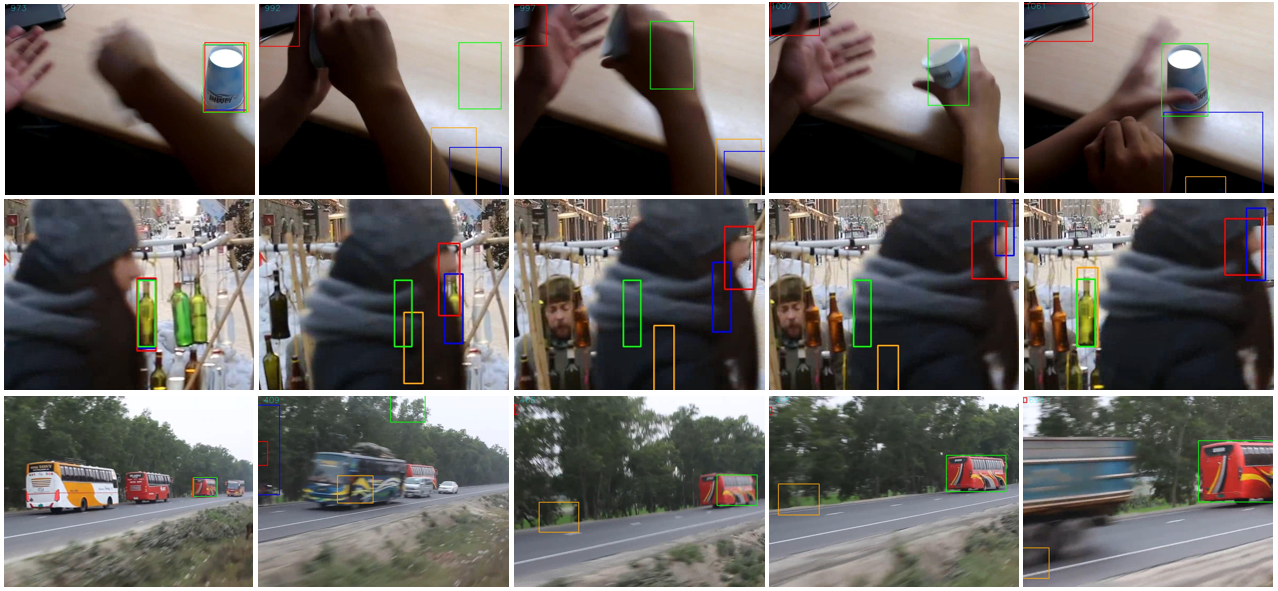}}
\caption{Illustration of three sequences in which UHP-SOT++ performs the best.
The tracking boxes of SiamRPN++, UHP-SOT++, ECO and ECO-HC are shown in
red, green, blue and yellow, respectively.} \label{fig:ourgoodcase}
\end{figure*}

\section{Exemplary Sequences and Qualitative Analysis}\label{sec:discussion}

After providing quantitative results in Sec. \ref{sec:experiments}, we
conduct error analysis on a couple of representative sequences to gain
more insights in this section.  Several exemplary sequences from LaSOT
are shown in Fig.~\ref{fig:error_showcase}, in which SiamRPN++ performs
either very well or quite poorly. In the first two sequences, we see
the power of accurate box regression contributed by the RPN. In this
type of sequences, good trackers can follow the object well. Yet, their
poor bounding boxes lead to a low success score. Furthermore, the
appearance model would be contaminated by the background information as
shown in the second cat example. The appearance model of DCF-based
methods learns background texture (rather than follows the cat)
gradually. When the box only covers part of the object, it might also
miss some object features, resulting in a degraded appearance model. In
both scenarios, the long-term performance will drop rapidly. Although
UHP-SOT++ allows the aspect ratio change to some extent as seen in the
first flag example, its residual map obtained by background motion
modeling is still not as effective as the RPN due to lack of semantic
meaning.  Generally speaking, the performance of UHP-SOT++ relies on the
quality of the appearance model and the residual map. 

On the other hand, SiamRPN++ is not robust enough to handle a wide range
of sequences well.  The third example sequence is from
video games. SiamRPN++ somehow includes background objects in its box
proposals and drifts away from its targets in the presented frames.
Actually, these background objects are different from their
corresponding target objects in either semantic meaning or local
information such as color or texture. The performance of the other three
trackers is not affected. We see that they follow the ground truth
without any problem. One explanation is that these video game sequences
could be few in the training set and, as a result, SiamRPN++ cannot
offer a reliable tracking result for them. 

Finally, several sequences in which UHP-SOT++ has the top performance
are shown in Fig. \ref{fig:ourgoodcase}. In the first cup sequence, all
other benchmarking methods lose the target while UHP-SOT++ could go back
to the object once the object has obvious motion in the scene. In the
second bottle sequence, UHP-SOT++ successfully detects occlusion without
making random guesses and the object box trajectory avoids the box to
drift away. In contrast, other trackers make ambitious moves without
considering the inertia of motion. The third bus sequence is a
complicated one that involves several challenges such as full occlusion,
scale change and aspect ratio change. UHP-SOT++ is the only one that can
recover from tracking loss and provide flexible box predictions. These
examples demonstrate the potential of UHP-SOT++ that exploits object and
background motion clues across frames effectively. 

\section{Conclusion and Future Work}\label{sec:conclusion}

An unsupervised high-performance tracker, UHP-SOT++, was proposed in
this paper.  It incorporated two new modules in the STRCF tracker
module. They were the background motion modeling module and the object
box trajectory modeling module.  Furthermore, a novel fusion strategy
was adopted to combine proposals from all three modules systematically.
It was shown by extensive experimental results on large-scale datasets
that UHP-SOT++ can generate robust and flexible object bounding boxes
and offer a real-time high-performance tracking solution on
resource-limited platforms. 

The pros and cons of supervised and unsupervised trackers were
discussed. Unsupervised trackers such as UHP-SOT and UHP-SOT++ have the
potential in delivering an explainable lightweight tracking solution
while maintaining good performance in accuracy. Supervised trackers such
as SiamRPN++ benefit from offline end-to-end learning and perform well
in general. However, they need to run on GPUs, which is too costly for
mobile and edge devices. They may encounter problems in rare samples.
Extensive supervision with annotated object boxes is costly.  Lack of
interpretability could be a barrier for further performance boosting. 

Although UHP-SOT++ offers a state-of-the-art unsupervised tracking
solution, there is still a performance gap between UHP-SOT++ and
SiamRPN++. It is worthwhile to find innovative ways to narrow down the
performance gap while keeping its attractive features such as
interpretability, unsupervised real-time tracking capability on small
devices, etc. as future extension. One idea is to find salient points in
the predicted and reference frames individually, develop a way to build
their correspondences, and reconstruct the object box in the predicted
frame based on its salient points. 



\bibliographystyle{IEEEtran}
\bibliography{ref}

\begin{thebibliography}{10}
\providecommand{\url}[1]{#1}
\csname url@samestyle\endcsname
\providecommand{\newblock}{\relax}
\providecommand{\bibinfo}[2]{#2}
\providecommand{\BIBentrySTDinterwordspacing}{\spaceskip=0pt\relax}
\providecommand{\BIBentryALTinterwordstretchfactor}{4}
\providecommand{\BIBentryALTinterwordspacing}{\spaceskip=\fontdimen2\font plus
\BIBentryALTinterwordstretchfactor\fontdimen3\font minus
  \fontdimen4\font\relax}
\providecommand{\BIBforeignlanguage}[2]{{%
\expandafter\ifx\csname l@#1\endcsname\relax
\typeout{** WARNING: IEEEtran.bst: No hyphenation pattern has been}%
\typeout{** loaded for the language `#1'. Using the pattern for}%
\typeout{** the default language instead.}%
\else
\language=\csname l@#1\endcsname
\fi
#2}}
\providecommand{\BIBdecl}{\relax}
\BIBdecl

\bibitem{xing2010multiple}
J.~Xing, H.~Ai, and S.~Lao, ``Multiple human tracking based on multi-view
  upper-body detection and discriminative learning,'' in \emph{2010 20th
  International Conference on Pattern Recognition}.\hskip 1em plus 0.5em minus
  0.4em\relax IEEE, 2010, pp. 1698--1701.

\bibitem{janai2020computer}
J.~Janai, F.~G{\"u}ney, A.~Behl, A.~Geiger \emph{et~al.}, ``Computer vision for
  autonomous vehicles: Problems, datasets and state of the art,''
  \emph{Foundations and Trends{\textregistered} in Computer Graphics and
  Vision}, vol.~12, no. 1--3, pp. 1--308, 2020.

\bibitem{zhang2015good}
G.~Zhang and P.~A. Vela, ``Good features to track for visual slam,'' in
  \emph{Proceedings of the IEEE conference on computer vision and pattern
  recognition}, 2015, pp. 1373--1382.

\bibitem{yilmaz2006object}
A.~Yilmaz, O.~Javed, and M.~Shah, ``Object tracking: A survey,'' \emph{Acm
  computing surveys (CSUR)}, vol.~38, no.~4, pp. 13--es, 2006.

\bibitem{fiaz2019handcrafted}
M.~Fiaz, A.~Mahmood, S.~Javed, and S.~K. Jung, ``Handcrafted and deep trackers:
  Recent visual object tracking approaches and trends,'' \emph{ACM Computing
  Surveys (CSUR)}, vol.~52, no.~2, pp. 1--44, 2019.

\bibitem{krizhevsky2012imagenet}
A.~Krizhevsky, I.~Sutskever, and G.~E. Hinton, ``Imagenet classification with
  deep convolutional neural networks,'' \emph{Advances in neural information
  processing systems}, vol.~25, pp. 1097--1105, 2012.

\bibitem{chatfield2014return}
K.~Chatfield, K.~Simonyan, A.~Vedaldi, and A.~Zisserman, ``Return of the devil
  in the details: Delving deep into convolutional nets,'' \emph{arXiv preprint
  arXiv:1405.3531}, 2014.

\bibitem{danelljan2017eco}
M.~Danelljan, G.~Bhat, F.~Shahbaz~Khan, and M.~Felsberg, ``Eco: Efficient
  convolution operators for tracking,'' in \emph{Proceedings of the IEEE
  conference on computer vision and pattern recognition}, 2017, pp. 6638--6646.

\bibitem{danelljan2016beyond}
M.~Danelljan, A.~Robinson, F.~S. Khan, and M.~Felsberg, ``Beyond correlation
  filters: Learning continuous convolution operators for visual tracking,'' in
  \emph{European conference on computer vision}.\hskip 1em plus 0.5em minus
  0.4em\relax Springer, 2016, pp. 472--488.

\bibitem{ma2015hierarchical}
C.~Ma, J.-B. Huang, X.~Yang, and M.-H. Yang, ``Hierarchical convolutional
  features for visual tracking,'' in \emph{Proceedings of the IEEE
  international conference on computer vision}, 2015, pp. 3074--3082.

\bibitem{qi2016hedged}
Y.~Qi, S.~Zhang, L.~Qin, H.~Yao, Q.~Huang, J.~Lim, and M.-H. Yang, ``Hedged
  deep tracking,'' in \emph{Proceedings of the IEEE conference on computer
  vision and pattern recognition}, 2016, pp. 4303--4311.

\bibitem{wang2018multi}
N.~Wang, W.~Zhou, Q.~Tian, R.~Hong, M.~Wang, and H.~Li, ``Multi-cue correlation
  filters for robust visual tracking,'' in \emph{Proceedings of the IEEE
  conference on computer vision and pattern recognition}, 2018, pp. 4844--4853.

\bibitem{li2018high}
B.~Li, J.~Yan, W.~Wu, Z.~Zhu, and X.~Hu, ``High performance visual tracking
  with siamese region proposal network,'' in \emph{Proceedings of the IEEE
  conference on computer vision and pattern recognition}, 2018, pp. 8971--8980.

\bibitem{li2019siamrpn++}
B.~Li, W.~Wu, Q.~Wang, F.~Zhang, J.~Xing, and J.~Yan, ``Siamrpn++: Evolution of
  siamese visual tracking with very deep networks,'' in \emph{Proceedings of
  the IEEE/CVF Conference on Computer Vision and Pattern Recognition}, 2019,
  pp. 4282--4291.

\bibitem{lu2018deep}
X.~Lu, C.~Ma, B.~Ni, X.~Yang, I.~Reid, and M.-H. Yang, ``Deep regression
  tracking with shrinkage loss,'' in \emph{Proceedings of the European
  conference on computer vision (ECCV)}, 2018, pp. 353--369.

\bibitem{nam2016learning}
H.~Nam and B.~Han, ``Learning multi-domain convolutional neural networks for
  visual tracking,'' in \emph{Proceedings of the IEEE conference on computer
  vision and pattern recognition}, 2016, pp. 4293--4302.

\bibitem{pu2018deep}
S.~Pu, Y.~Song, C.~Ma, H.~Zhang, and M.-H. Yang, ``Deep attentive tracking via
  reciprocative learning,'' \emph{arXiv preprint arXiv:1810.03851}, 2018.

\bibitem{song2017crest}
Y.~Song, C.~Ma, L.~Gong, J.~Zhang, R.~W. Lau, and M.-H. Yang, ``Crest:
  Convolutional residual learning for visual tracking,'' in \emph{Proceedings
  of the IEEE international conference on computer vision}, 2017, pp.
  2555--2564.

\bibitem{bertinetto2016fully}
L.~Bertinetto, J.~Valmadre, J.~F. Henriques, A.~Vedaldi, and P.~H. Torr,
  ``Fully-convolutional siamese networks for object tracking,'' in
  \emph{European conference on computer vision}.\hskip 1em plus 0.5em minus
  0.4em\relax Springer, 2016, pp. 850--865.

\bibitem{tao2016siamese}
R.~Tao, E.~Gavves, and A.~W. Smeulders, ``Siamese instance search for
  tracking,'' in \emph{Proceedings of the IEEE conference on computer vision
  and pattern recognition}, 2016, pp. 1420--1429.

\bibitem{zhu2018distractor}
Z.~Zhu, Q.~Wang, B.~Li, W.~Wu, J.~Yan, and W.~Hu, ``Distractor-aware siamese
  networks for visual object tracking,'' in \emph{Proceedings of the European
  Conference on Computer Vision (ECCV)}, 2018, pp. 101--117.

\bibitem{wang2018learning}
Q.~Wang, Z.~Teng, J.~Xing, J.~Gao, W.~Hu, and S.~Maybank, ``Learning
  attentions: residual attentional siamese network for high performance online
  visual tracking,'' in \emph{Proceedings of the IEEE conference on computer
  vision and pattern recognition}, 2018, pp. 4854--4863.

\bibitem{he2018twofold}
A.~He, C.~Luo, X.~Tian, and W.~Zeng, ``A twofold siamese network for real-time
  object tracking,'' in \emph{Proceedings of the IEEE Conference on Computer
  Vision and Pattern Recognition}, 2018, pp. 4834--4843.

\bibitem{wang2021transformer}
N.~Wang, W.~Zhou, J.~Wang, and H.~Li, ``Transformer meets tracker: Exploiting
  temporal context for robust visual tracking,'' in \emph{Proceedings of the
  IEEE/CVF Conference on Computer Vision and Pattern Recognition}, 2021, pp.
  1571--1580.

\bibitem{chen2021transformer}
X.~Chen, B.~Yan, J.~Zhu, D.~Wang, X.~Yang, and H.~Lu, ``Transformer tracking,''
  in \emph{Proceedings of the IEEE/CVF Conference on Computer Vision and
  Pattern Recognition}, 2021, pp. 8126--8135.

\bibitem{bolme2010visual}
D.~S. Bolme, J.~R. Beveridge, B.~A. Draper, and Y.~M. Lui, ``Visual object
  tracking using adaptive correlation filters,'' in \emph{2010 IEEE computer
  society conference on computer vision and pattern recognition}.\hskip 1em
  plus 0.5em minus 0.4em\relax IEEE, 2010, pp. 2544--2550.

\bibitem{henriques2014high}
J.~F. Henriques, R.~Caseiro, P.~Martins, and J.~Batista, ``High-speed tracking
  with kernelized correlation filters,'' \emph{IEEE transactions on pattern
  analysis and machine intelligence}, vol.~37, no.~3, pp. 583--596, 2014.

\bibitem{danelljan2015convolutional}
M.~Danelljan, G.~Hager, F.~Shahbaz~Khan, and M.~Felsberg, ``Convolutional
  features for correlation filter based visual tracking,'' in \emph{Proceedings
  of the IEEE international conference on computer vision workshops}, 2015, pp.
  58--66.

\bibitem{danelljan2016discriminative}
M.~Danelljan, G.~H{\"a}ger, F.~S. Khan, and M.~Felsberg, ``Discriminative scale
  space tracking,'' \emph{IEEE transactions on pattern analysis and machine
  intelligence}, vol.~39, no.~8, pp. 1561--1575, 2016.

\bibitem{bertinetto2016staple}
L.~Bertinetto, J.~Valmadre, S.~Golodetz, O.~Miksik, and P.~H. Torr, ``Staple:
  Complementary learners for real-time tracking,'' in \emph{Proceedings of the
  IEEE conference on computer vision and pattern recognition}, 2016, pp.
  1401--1409.

\bibitem{valmadre2017end}
J.~Valmadre, L.~Bertinetto, J.~Henriques, A.~Vedaldi, and P.~H. Torr,
  ``End-to-end representation learning for correlation filter based tracking,''
  in \emph{Proceedings of the IEEE conference on computer vision and pattern
  recognition}, 2017, pp. 2805--2813.

\bibitem{li2018learning}
F.~Li, C.~Tian, W.~Zuo, L.~Zhang, and M.-H. Yang, ``Learning spatial-temporal
  regularized correlation filters for visual tracking,'' in \emph{Proceedings
  of the IEEE conference on computer vision and pattern recognition}, 2018, pp.
  4904--4913.

\bibitem{danelljan2015learning}
M.~Danelljan, G.~Hager, F.~Shahbaz~Khan, and M.~Felsberg, ``Learning spatially
  regularized correlation filters for visual tracking,'' in \emph{Proceedings
  of the IEEE international conference on computer vision}, 2015, pp.
  4310--4318.

\bibitem{wang2019unsupervised}
N.~Wang, Y.~Song, C.~Ma, W.~Zhou, W.~Liu, and H.~Li, ``Unsupervised deep
  tracking,'' in \emph{Proceedings of the IEEE/CVF Conference on Computer
  Vision and Pattern Recognition}, 2019, pp. 1308--1317.

\bibitem{wang2021unsupervised}
N.~Wang, W.~Zhou, Y.~Song, C.~Ma, W.~Liu, and H.~Li, ``Unsupervised deep
  representation learning for real-time tracking,'' \emph{International Journal
  of Computer Vision}, vol. 129, no.~2, pp. 400--418, 2021.

\bibitem{wu2021progressive}
Q.~Wu, J.~Wan, and A.~B. Chan, ``Progressive unsupervised learning for visual
  object tracking,'' in \emph{Proceedings of the IEEE/CVF Conference on
  Computer Vision and Pattern Recognition}, 2021, pp. 2993--3002.

\bibitem{zhou2021uhp}
Z.~Zhou, H.~Fu, S.~You, C.~C. Borel-Donohue, and C.-C.~J. Kuo, ``Uhp-sot: An
  unsupervised high-performance single object tracker,'' \emph{arXiv preprint
  arXiv:2110.01812}, 2021.

\bibitem{7001050}
Y.~Wu, J.~Lim, and M.-H. Yang, ``Object tracking benchmark,'' \emph{IEEE
  Transactions on Pattern Analysis and Machine Intelligence}, vol.~37, no.~9,
  pp. 1834--1848, 2015.

\bibitem{danelljan2014adaptive}
M.~Danelljan, F.~Shahbaz~Khan, M.~Felsberg, and J.~Van~de Weijer, ``Adaptive
  color attributes for real-time visual tracking,'' in \emph{Proceedings of the
  IEEE Conference on Computer Vision and Pattern Recognition}, 2014, pp.
  1090--1097.

\bibitem{hariharakrishnan2005fast}
K.~Hariharakrishnan and D.~Schonfeld, ``Fast object tracking using adaptive
  block matching,'' \emph{IEEE transactions on multimedia}, vol.~7, no.~5, pp.
  853--859, 2005.

\bibitem{aggarwal2006object}
A.~Aggarwal, S.~Biswas, S.~Singh, S.~Sural, and A.~K. Majumdar, ``Object
  tracking using background subtraction and motion estimation in mpeg videos,''
  in \emph{Asian Conference on Computer Vision}.\hskip 1em plus 0.5em minus
  0.4em\relax Springer, 2006, pp. 121--130.

\bibitem{liang2015encoding}
P.~Liang, E.~Blasch, and H.~Ling, ``Encoding color information for visual
  tracking: Algorithms and benchmark,'' \emph{IEEE Transactions on Image
  Processing}, vol.~24, no.~12, pp. 5630--5644, 2015.

\bibitem{mueller2016benchmark}
M.~Mueller, N.~Smith, and B.~Ghanem, ``A benchmark and simulator for uav
  tracking,'' in \emph{European conference on computer vision}.\hskip 1em plus
  0.5em minus 0.4em\relax Springer, 2016, pp. 445--461.

\bibitem{fan2019lasot}
H.~Fan, L.~Lin, F.~Yang, P.~Chu, G.~Deng, S.~Yu, H.~Bai, Y.~Xu, C.~Liao, and
  H.~Ling, ``Lasot: A high-quality benchmark for large-scale single object
  tracking,'' in \emph{Proceedings of the IEEE/CVF Conference on Computer
  Vision and Pattern Recognition}, 2019, pp. 5374--5383.

\bibitem{alan2018discriminative}
L.~Alan, T.~Voj{\'\i}{\v{r}}, L.~{\v{C}}ehovin, J.~Matas, and M.~Kristan,
  ``Discriminative correlation filter tracker with channel and spatial
  reliability,'' \emph{International Journal of Computer Vision}, vol. 126,
  no.~7, pp. 671--688, 2018.

\bibitem{ma2015long}
C.~Ma, X.~Yang, C.~Zhang, and M.-H. Yang, ``Long-term correlation tracking,''
  in \emph{Proceedings of the IEEE conference on computer vision and pattern
  recognition}, 2015, pp. 5388--5396.

\end{thebibliography}

\end{document}